\documentclass{bmvc2k}

\title{Self-supervised Learning for Video Correspondence Flow}
\addauthor{Zihang Lai}{zihang.lai@cs.ox.ac.uk}{1}
\addauthor{Weidi Xie}{weidi@robots.ox.ac.uk}{2}

\addinstitution{
Department of Computer Science\\
University of Oxford, UK
}

\addinstitution{
 Visual Geometry Group,\\
 Department of Engineering Science\\
 University of Oxford, UK
}

\runninghead{Lai, Xie}{Self-supervised Learning for Video Correspondence Flow}

\usepackage[title]{appendix}

\usepackage{graphicx}
\usepackage{graphics} % for pdf, bitmapped graphics files
\usepackage[normalem]{ulem}
\usepackage{array}
\usepackage{hyperref}
\useunder{\uline}{\ul}{}
\usepackage{multirow}
\usepackage{pifont}
\usepackage{sidecap}
\usepackage{booktabs}
\usepackage{floatrow}
\usepackage{amsfonts}

\newfloatcommand{capbtabbox}{table}[][\FBwidth]

\newcommand{\cmark}{\ding{51}}%
\newcommand{\xmark}{\ding{55}}%

\def\ie{\emph{i.e}\bmvaOneDot}
\def\eg{\emph{e.g}\bmvaOneDot}

\def\etal{\emph{et al}\bmvaOneDot}

\begin{document}

\maketitle

\begin{abstract}
The objective of this paper is \emph{self-supervised learning} of feature embeddings 
that are suitable for matching correspondences along the videos, which we term correspondence flow.
By leveraging the natural spatial-temporal coherence in videos, 
we propose to train a ``pointer'' that reconstructs a target frame by copying pixels from a reference frame.

We make the following contributions:
\emph{First}, 
we introduce a simple information bottleneck that forces the model to learn robust features for correspondence matching, 
and prevent it from learning trivial solutions, \eg matching based on low-level colour information.
\emph{Second},
to alleveate tracker drifting, due to complex object deformations, illumination changes and occlusions,
we propose to train a recursive model over long temporal windows with scheduled sampling and cycle consistency.
\emph{Third},
we achieve the state-of-the-art performance on DAVIS 2017 video segmentation and JHMDB keypoint tracking tasks,
outperforming all previous self-supervised learning approaches by a significant margin.
\emph{Fourth},
in order to shed light on the potential of self-supervised learning on the task of video correspondence flow,
we probe the upper bound by training on additional data, \ie more diverse videos,
further demonstrating significant improvements on video segmentation.
The source code will be released at \url{https://github.com/zlai0/CorrFlow}.
\end{abstract} 

%!TEX root=../root.tex

\section{Introduction}
Correspondence matching is a fundamental building block for numerous applications ranging from depth estimation~\cite{Kendall17}
and optical flow~\cite{Brox04,Dosovitskiy15,Ilg17}, 
to segmentation and tracking~\cite{Smith95b}, and 3D reconstruction~\cite{Hartley2004}. 
However, training models for correspondence matching is not trivial, 
as obtaining manual annotations can be prohibitively expensive, 
and sometimes is even impossible due to occlusions and complex object deformations.
In the recent works~\cite{Rocco17,Rocco18}, 
Rocco~\etal proposed to circumvent this issue by pre-training Convolutional Neural Networks~(CNNs) for predicting artificial transformations,
and further bootstrap the model by finetuning on a small dataset with human annotations.
Alternatively, 
the necessity for labbelled data can be avoided by using self-supervised learning,
\ie a form of unsupervised learning, where part of the data is withheld for defining a proxy task,
such that the model will be forced to learn the semantic representation that we really care about.

Videos have shown to be appealing as a data source for self-supervised learning
due to their almost infinite supply~(from YouTube etc),
and the availability of numerous proxy losses that can be employed from the intrinsic spatio-temporal coherence,
\ie the signals in video tend to vary smoothly in time~\cite{Wiskott02,Jayaraman16}. 
In this paper, we propose to tackle the task of correspondence flow in videos with self-supervised learning. 
We are certainly not the first to explore this idea,
in the seminal paper by Vondrick~\etal~\cite{Vondrick18}, 
they propose to learn embeddings for grayscale frames and use \emph{colorization} as a proxy task for training.
Despite the promising efforts, 
the model capacity has been significantly constrained due to the loss of colour information,
and suffers the problem of tracker drifting as only pair of frames are used during training.

% ==============
\begin{figure}[t]
\begin{center}
\setlength{\tabcolsep}{1pt}
\begin{tabular}{cc}
\includegraphics[width=0.592\textwidth]{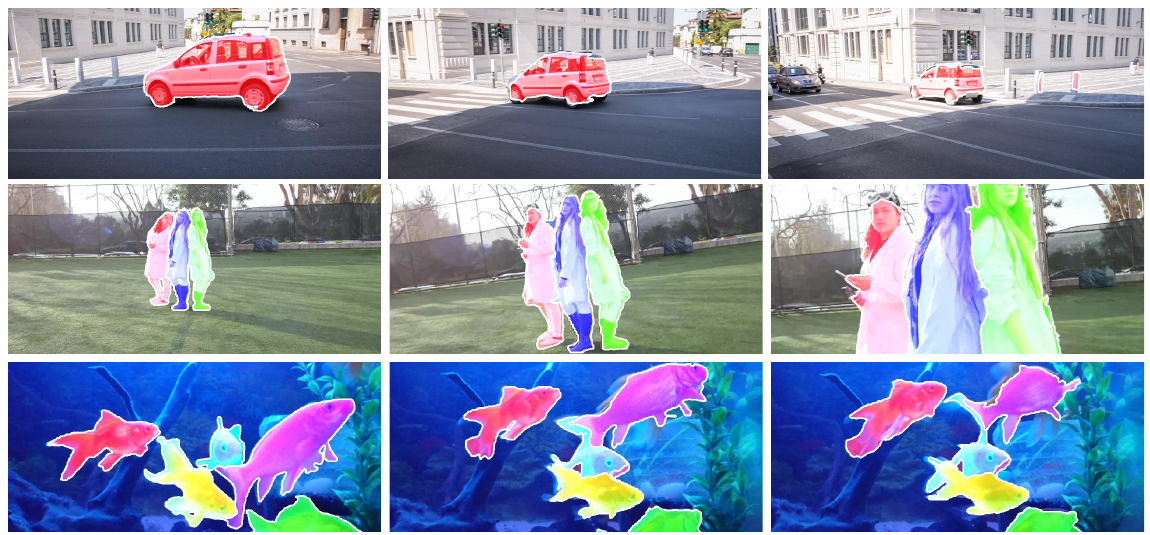} &
\includegraphics[width=0.365\textwidth]{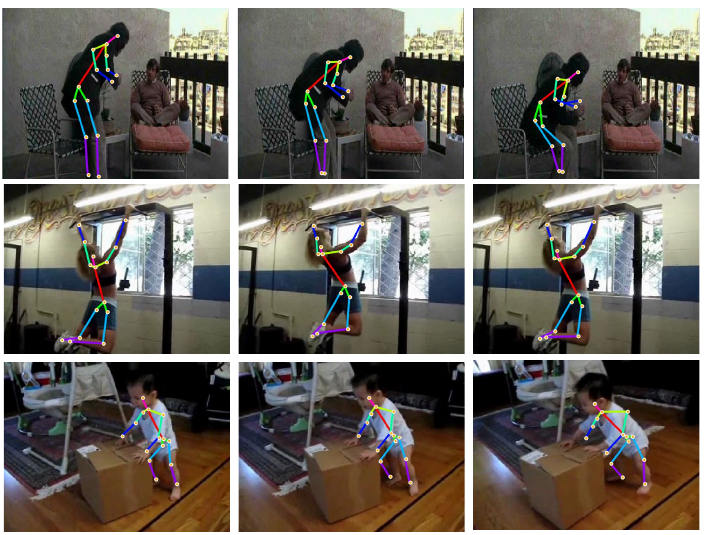} \\
\multicolumn{1}{c}{(a) DAVIS 2017 Video Segmentation} &  
\multicolumn{1}{c}{(b) Keypoint Tracking} 
\end{tabular}
\vspace{-20pt}
\caption[]{
We propose self-supervised learning of correspondence flow on videos.
Without any fine-tuning, the acquired representation generalizes to various tasks: 
(a) video segmentaiton;  (b) keypoint tracking.
\textbf{Note}: For both tasks, the annotation for the first frame is given, the goal is to propagate the annotations through the videos.}
\label{fig:qualitative}
\end{center}
\end{figure}

We make the following contributions:
\emph{First}, 
we introduce an embarrassingly simple idea to avoid trivial solutions while learning pixelwise matching by frame reconstruction.
During training, channel-wise dropout and colour jitterings are added intentionally on the input frames,
the model is therefore forced \emph{not} to rely on low-level colour information, and must be robust to colour jittering.
\emph{Second},
we propose to train the model recursively on videos over long temporal windows with scheduled sampling and forward-backward consistency.
%In the training stage, later frames may not have access to the ground truth previous frame, 
%but be reconstructed by copying pixels from the previous predictions.
Both ideas have shown to improve the model robustness and help to alleviate the  tracker drifting problem.
\emph{Third},
after self-supervised training,
we benchmark the model on two downstream tasks focusing on pixelwise tracking, 
\eg DAVIS 2017 video segmentation and JHMDB keypoint tracking, 
outperforming all previous self-supervised learning approaches by a significant margin. 
\emph{Fourth},
to further shed light on the potential of self-supervised learning for video correspondence flow,
we probe the upper bound by training on more diverse video data,
and further demonstrating significant improvements on video segmentation.

%During training,
%a pair of images are fed to a siamese feature encoder, 
%and an affinity matrix is computed which copies colors from the reference frame.
%To supervise the training, a pixel-wise loss is applied between the reconstructed and ground truth target frame.
%In another recent work by Wang~\etal~\cite{Wang19}, 
%the authors propose to use cycle-consistency as the supervision signal,
%and train an embedding for tracking patches along in a video clip.
%Our paper is of direct relevance to Vondrick~\etal~\cite{Vondrick18}.

%Moreover,
%we notice that training the model on Kinetics is not ideal as it is human-centric video dataset, 
%most of the classes in DAVIS are not covered in Kinetics, \eg animals.
%Therefore, there is no way for the model to learn semantics matching for these unseen classes.
%As proof of concept, 
%we probe $8$ different classes in DAVIS by downloading videos from YouTube 
%and finetuning the model on them in a self-supervised way.
%Note that, we only download videos by the class labels, 
%no segmentation annotations are used while finetuning.
%After self-supervised finetuning, 
%the model shows a significant improvement on video segmentation.

%!TEX root=root.tex
\vspace{-10pt}
\section{Related Work}
\par \noindent \textbf{Correspondence Matching.}
Recently, 
many researchers have studied the task of correspondence matching using deep Convolutional Neural Networks~\cite{Han17,Kim17,Novotny17,Rocco17, Rocco18}.
The works from Rocco~\etal~\cite{Rocco17, Rocco18}, 
propose to train CNNs by learning the artificial transformations between pairs of images.
For robust estimation, 
they applied a differentiable soft inlier score for evaluating the quality of alignment 
between spatial features and providing a loss for learning semantic correspondences. 
However, their work may not be ideal as the model still relies on synthetic transformations.
In contrast, we address the challenge of learning correspondence matching by exploiting the temporal coherence in videos.

\vspace{2pt}
\par \noindent \textbf{Optical Flow.}
In the conventional variational approaches, 
optical flow estimation is treated as an energy minimization problem based on brightness constancy and spatial smoothness~\cite{Horn81}.
In later works, feature matching is used to firstly establish sparse matchings, 
and then interpolated into dense flow maps in a pyramidal coarse-to-fine manner~\cite{Brox09,Revaud15,Weinzaepfel13}. 
Recently,
convolutional neural networks (CNNs) have been applied to improve the matching by learning effective feature embeddings~\cite{Bailer17,Jia17}.
Another line of more relevant research is unsupervised learning for optical flow.
The basic principles are based on brightness constancy and spatial smoothness~\cite{Yu16,Wang18,Meister18}. 
This leads to the popular photometric loss which measures the difference between the reference image and the warped image.
For occluded regions, a mask is implicitly estimated by checking forward-backward consistency.

\vspace{2pt}
\par \noindent \textbf{Self-supervised Learning.}
Of more direct relevance to our training framework are self-supervised frameworks that use video data~\cite{Agrawal15,Denton17,Fernando17, Gan18,Jakab18,Jia16, Lee17, Misra16,Wang15,Wiles18, Isola15,Jayaraman16,Jayaraman15,Jing18,Kim18,Vondrick18}.
In~\cite{Misra16,Fernando17,Wei18}, the proxy task is defined to focus on temporal sequence ordering of the frames.
Another approach is to use the temporal coherence as a proxy loss~\cite{Isola15,Jayaraman16,Wang15}.
Other approaches use egomotion~\cite{Agrawal15,Jayaraman15} in order
to enforce equivariance in feature space \cite{Jayaraman15}. 
Recently~\cite{Vondrick18}, leveraged  the natural temporal coherency of colour in videos, to train
a network for tracking and correspondence related tasks.
Our approach builds in particular on those that use frame synthesis~\cite{Denton17,Jia16,Wiles18}, 
though for us synthesis is a proxy task rather than the end goal.

%!TEX root=../root.tex
% ---------------------
\vspace{-10pt}
\section{Approach}
The goal of this work is to train an embedding network with self-supervised learning that enables pixelwise correspondence matching.
Our basic idea is to exploit spatial-temporal coherence in videos,
that is, the frame appearances will not change abruptly,
and colours can act as a reliable supervision signal for learning correspondences.

% ---------------------
\vspace{-6pt}
\subsection{Background}
% ---------------------
\label{subsec:background}
In this section, we briefly review the work by Vondrick~\etal~\cite{Vondrick18}.
Formally, let $c_i \in \mathbb{R}^d$ be the true colour for pixel $i$ in the reference frame, 
and let $c_j \in \mathbb{R}^d$ be the true colour for a pixel $j$ in the target frame.
$y_j \in \mathbb{R}^d$ is the model's prediction for $c_j$, 
it is a linear combination of colours in the reference frame:
\begin{align}
y_j = \sum_i A_{ij}c_i , \text{\hspace{15pt} where }
A_{ij} = \frac{exp\langle f_i^T f_j \rangle }{\sum_k exp \langle f_k^T f_j \rangle}
\end{align}
$A$ is an affinity matrix computed from simple dot products
between the feature embeddings of the \emph{grayscale} target and reference frame~($f$'s).

Despite the efforts in circumventing trivial solutions, 
training models with \emph{grayscale} inputs has introduces a train-test discrepancy when deploying to the downstream task,
as the model has never been trained to encode the correlation of RGB channels.
Another issule lies in the fact that their model is only trained with pairs of ground truth video frames,
which inevitably leads to model drifting when evaluated on video tracking tasks.
As the prediction for later steps rely on the prediction from previous steps,
the errors accumulate,
and there is no mechanism for the model to recover from previous error states.

In the following sections, 
we propose to train a framework that aims to close the gap between training and testing as much as possible,
\ie the model should ideally be trained on \emph{full-colour} and \emph{high-resolution} over \emph{long} video sequences.

% ---------------------
%         Figure
% ---------------------
\begin{figure*}[t]
\begin{center}
\includegraphics[width=\textwidth]{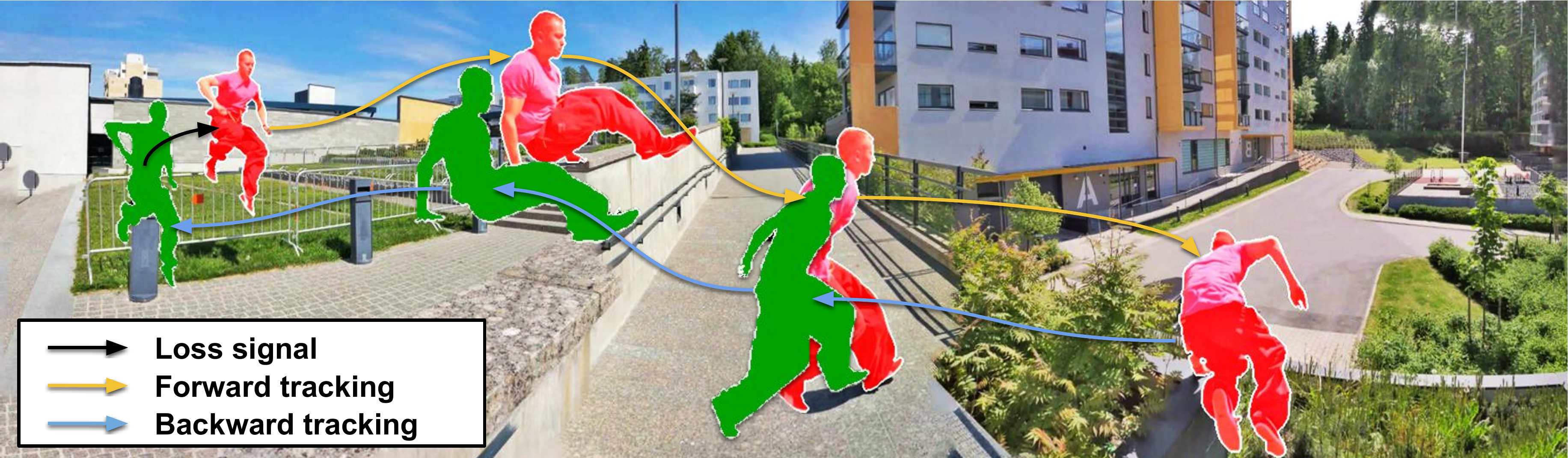} 
\vspace{-25pt}
\caption[]{
\small{An overview of the proposed self-supervised learning for correspondence flow.
A recursive model is used to compute the dense correspondence matching over a long temporal window 
with forward-backward cycle consistency.}}
\label{fig:layout}
\end{center}
\end{figure*}

% ---------------------
\vspace{-6pt}
\subsection{Feature Embedding with Information Bottleneck}
\label{subsec:feature_enc}
% ---------------------
%We are able to train our model using full colour images through applying several ``jitterings'' to the input image.
Given a collection of frames $\{I_1, I_2, ..., I_N\}$ from a video clip, we parametrize the feature embedding module with CNNs: 
\begin{align}
f_i = \Phi(g(I_i); \theta)
\end{align}
where $\Phi$ refers to a ResNet feature encoder~(details in Arxiv paper\footnote{\url{https://arxiv.org/abs/1905.00875}}), 
and $g(\cdot)$ denotes an \emph{information bottleneck} that prevents from information leakage.
In Vondrick~\etal~\cite{Vondrick18}, this is simply defined as a function that converts RGB frame into \emph{grayscale},
therefore, the model is forced \emph{not} to rely on colours for matching correspondences.

We propose to randomly zero out 0, 1, or 2 channels in each input frame 
with some probability~(one possible input is shown in Figure~\ref{fig:network}~(a)), 
and perturb the brightness, contrast and saturation of an image by up to $10\%$.
Despite the simplicity, this design poses two benefits:
\emph{First}, the input jitterings and stochastic dropout are essentially acting as an \emph{information bottleneck},
which prevents the model from co-adaptation of low-level colours.
When deploying for downstream tasks, 
images of full RGB colours are taken as input directly~(no jittering).
\emph{Second}, 
it can also be treated as data augmentation that potentially improves the model robustness under challenging cases,
for instance, illumination changes.

% ---------------------
\vspace{-6pt}
\subsection {Restricted Attention}
\label{subsec:restrict_att}
\begin{figure*}[t]
\begin{center}
\setlength{\tabcolsep}{1pt}
\begin{tabular}{cp{0.1cm}|p{0.1cm}cp{0.1cm}c}
\includegraphics[width=0.51\textwidth]{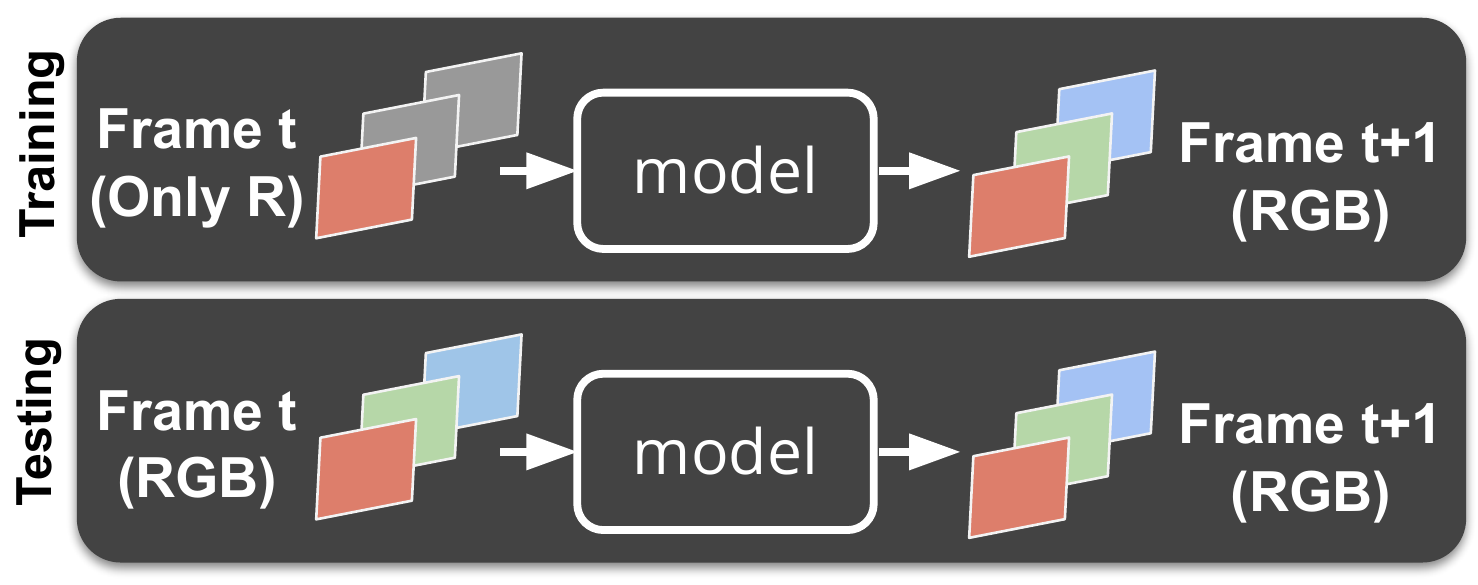} & & &
\includegraphics[width=0.2\textwidth]{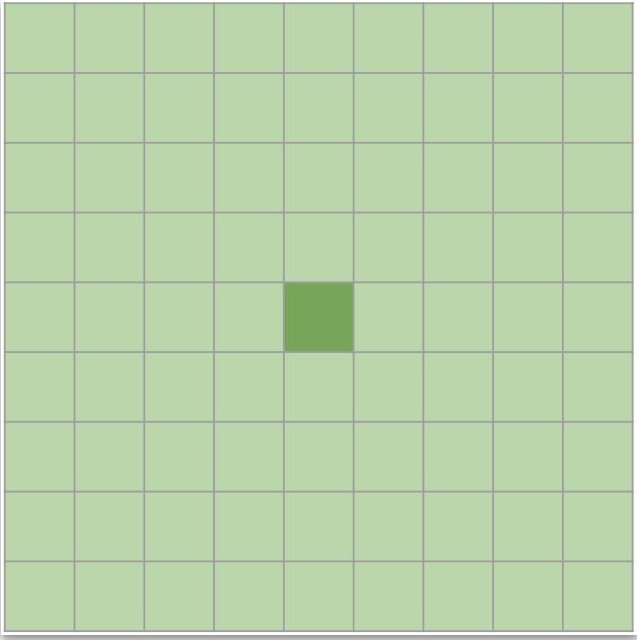} &  &
\includegraphics[width=0.2\textwidth]{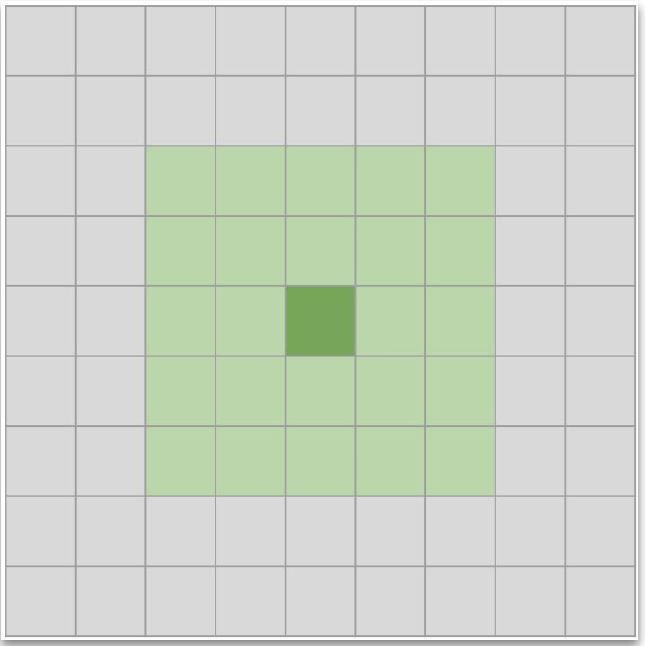} \\
\multicolumn{1}{c}{\textbf{\footnotesize (a) Colour Dropout}} & & &
\multicolumn{1}{c}{\textbf{\footnotesize (b) Full Attention} } &  &
\multicolumn{1}{c}{\textbf{\footnotesize (c) Restricted Attention}} \\
\end{tabular}
\vspace{-20pt}
\caption[]{
\small{Restricted attention and colour dropout. See text for details.}}
\label{fig:network}
\end{center}
\end{figure*}

%We are able to train our model using high-resolution training images by using restricted attention,
In the previous work~\cite{Vondrick18}, \emph{full attention} has been used for computing the affinity matrix,
\ie all pairs of pixels in target and reference frames are correlated~(Figure~\ref{fig:network}~(b)). 
However, the memory and computational consumption tends to grow quadratically with the spatial footprint of the feature maps, 
thus limiting the resolution.
In fact, videos are full of regularities, 
\ie the appearances in the video clip tend to change smoothly both in spatial and temporal axes.
To fully exploit this property,
we propose to use a restricted attention mechanism~(Figure~\ref{fig:network}~(c)), 
which leads to dramatic decrease in computation and memory consumption,
and enables to train on \emph{high-resolution} frames.

% ---------------------
%         Figure
% ---------------------

Specifically, we impose a maximum disparity of $M$, 
\ie pixels in the reference frame $t$ are searched for locally in a square patch of size $(2M+1)\times (2M+1)$, 
centered at the target pixel. 
Suppose the feature maps have a dimension of $H\times W$, 
the affinity volume~($A$) is therefore a $4D$ tensor of dimension $H \times W \times (2M+1) \times (2M+1)$. 
The $(i,j,k,l)$ entry of the tensor denotes the similarity between pixel $(i,j)$ of the target frame, 
and pixel $(i+k-M, j+l-M)$ of the reference frame.  
\begin{align}
A^{ijkl} &= \frac{\exp{\langle f_t^{(i+k-M)(j+l-M)}, 
f_{t+1}^{ij}\rangle} } {\sum_p\sum_q\exp{\langle f_t^{(i+q)(j+p)}, f_{t+1}^{ij}\rangle}} \\
\hat{I}_{t+1} &= \psi(A_{(t,t+1)}, I_t) = \sum_{p}\sum_q A^{ij(p+M)(q+M)}I_t
\end{align}
\noindent where $p,q \in [-M,M]$,
$f_t = \Phi(g(I_t);\theta)$ and  $f_{t+1} = \Phi(g(I_{t+1}); \theta)$
refer to the feature embeddings for frame $t$ and $t+1$ respectively.
$\psi()$ denotes a soft-copy operation for reconstructing $\hat{I}_{t+1}$ by ``borrowing'' colours from $I_t$ frame.

% ---------------------
\vspace{-6pt}
\subsection{Long-term Correspondence Flow}
\label{subsec:longterm}
% --------------------- 
One of the challenges on self-supervised learning of correspondence flow is how to sample the training frames;
If two frames are sampled closely in the temporal axis,
the objects remain unchanged in both appearance and spatial position,
matching becomes a trivial task and the model will not benefit from training on them.
In the contrary, if the frames are sampled with a large temporal stride,
the assumption of using reconstruction as supervision may fail,  
due to complex object deformation, illumination change, motion blurs, and occlusions.

In this section,
we propose two approaches to improve the model's robustness to tracker drifting, 
and gently bridge the gap of training with samples that are neither easy nor that difficult,
\ie scheduled sampling, and cycle consistency.

% ---------------------
\vspace{-6pt}
\subsubsection{Scheduled Sampling}
\label{subsec:schedule}
% ---------------------
Scheduled sampling is a widely used curriculum learning strategy for sequence-to-sequence models~\cite{Bengio15},
the main idea is to replace some ground truth tokens by the model's prediction, 
therefore improve the robustness and bridge the gap between train and inference stage.

In our case, for $n$ frames in a video sequence, 
a shared embedding network is used to get feature embeddings~($f_i = \Phi(g(I_i);\theta)$ where $i = 1,..., n$), 
the reconstruction is therefore formalized as a recursive process:
$$\hat{I}_n =
\begin{cases}
\psi(A_{(n-1,n)}, I_{n-1}) & \text{\small{(1)}}\\
\psi(A_{(n-1,n)}, \hat{I}_{n-1})& \text{\small{(2)}}
\end{cases}$$
while reconstructing the $n$th frame~($\hat{I}_n$), 
the model may have access to the previous frame as either groundtruth~($I_{n-1}$) or model prediction~($\hat{I}_{n-1}$).
During training, the probability of using ground truth frames starts from a higher value~($0.9$) in early training stage, 
and is uniformly annealed to a probability of $0.6$.
Note that, as the model is recursive,
the scheduled sampling forces the model to recover from error states and to be robust to drifting.

% ---------------------
\vspace{-6pt}
\subsubsection{Cycle Consistency}
\label{subsec:cycle}
% ---------------------
Following the scheduled sampling, 
we also explicitly adopt a cycle consistency for training correspondence flow. 
Unlike~\cite{Wang19},
we do not use the cycle consistency as the dominating supervision signal,
instead, it is treated as another regularizer for combating drifting.
During training, we apply our model $n$ frames forward and backward to the current frame.

% ---------------------
\vspace{-6pt}
\subsection{Training Objectives}
% ---------------------
Similar to~\cite{Vondrick18},
we pose frame reconstruction as a classification problem,
the colour for each pixel is quantized into $16$ classes with K-means clustering in the \textit{Lab} space.
The objective function is defined as :
\begin{align}
L = \alpha_1 \cdot \sum_{i=1}^{n} \mathcal{L}_1(I_i, \hat{I}_i) + \alpha_2 \cdot \sum_{j=n}^{1} \mathcal{L}_2(I_j, \hat{I}_j)
\end{align}
where $\mathcal{L}_1, \mathcal{L}_2$ refer to the pixel-wise cross entropy between groundtruth and reconstructed frames in the \emph{forward} and \emph{backward} paths, 
the loss weights for both paths are set as $\alpha_1 = 1.0$, $\alpha_2 = 0.1$ respectively,
\ie \emph{forward} path is weighted more than \emph{backward} path.

%!TEX root=../root.tex
\vspace{-6pt}
\section{Experiments and Analysis}
In the following sections, we start with the training details, followed by ablation studies,
\eg colour dropout, restricted attention, scheduled sampling and cycle consistency.

\vspace{3pt}
\par \noindent \textbf{Training Details}
In this paper, we train CNNs in a completely self-supervised manner on Kinetics~\cite{Kay17}, 
meaning we do \emph{not} use any information other than video sequences, and \emph{not} finetune for any target task. 
As pre-processing,
we decode the videos with a frame rate of $6$fps, and resize all frames to $256\times 256\times3$. 
In all of our experiments, 
we use a variant of ResNet-18 as a feature encoder
~(we refer the reader to our Arxiv paper for more details\footnote{\url{https://arxiv.org/abs/1905.00875}}).
which ends up with feature embeddings with spatial resolution of $1/4$ the original image.
The max disparity $M$ in the restricted attention is set to be $6$~(as described in Section~\ref{subsec:restrict_att}).
The temporal length $n$ is set to $3$ in our case, 
so when considering the forward-backward cycles, the sequence length is actually $5$ frames.
We train our model end-to-end using a batch size of $8$ for 1M iterations with an Adam optimizer. 
The initial learning rate is set to $2e^{-4}$, and halved on 0.4, 0.6 and 0.8M iterations. 

\vspace{3pt}
\par \noindent \textbf{Evaluation Metrics}
In this paper, we report results on two public benchmarks:
video segmentation, and pose keypoint propagation.
For both tasks,
a ground truth annotation is given for the first frame, 
and the objective is to propagate the mask to subsequent frames.
In video segmentation, we benchmark our model on DAVIS-2017~\cite{Pont-Tuset17}, 
two standard metrics are used, \ie region overlapping ($\mathcal{J}$) and contour accuracy ($\mathcal{F}$).
For pose keypoint tracking, we evaluate our model on JHMDB dataset and report two different PCK metrics. 
The first (PCK$_{instance}$) considers a keypoint to be correct if the 
Normalized Euclidean Distance between that keypoint and the ground truth is smaller than a threshold $\alpha$. 
The second (PCK$_{max}$) accepts a keypoint if it is located within $\alpha\cdot\max(w, h)$ pixels of the ground truth, 
where $w$ and $h$ are the width and height of the instance bounding box. 

% ---------------
\vspace{-6pt}
\subsection{Video Segmentation on DAVIS-2017}
% ---------------

\subsubsection{Ablation Studies}
To examine the effects of different components, 
we conduct a series of ablation studies by removing one component at a time.
All models are trained from scratch on Kinetics, 
and evaluated on the video segmentation task~(DAVIS-2017) without finetuning.

% --------------------
%         Figure
% --------------------
\begin{figure}[H]
\begin{floatrow}
\ffigbox{%
  \includegraphics[width=0.45\textwidth]{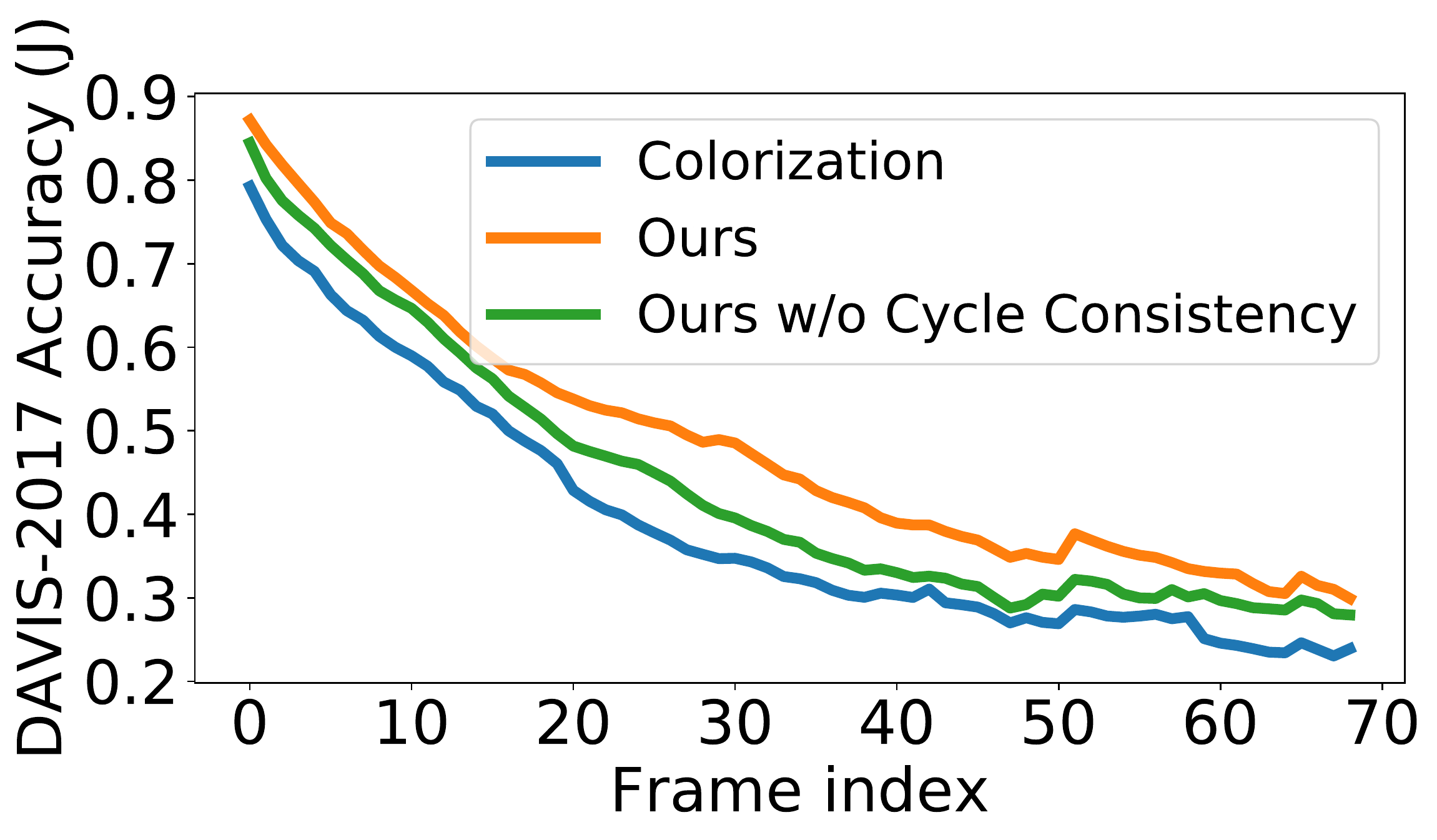}
}
{%
\vspace{-13pt}
\caption{Model comparison on the problem of tracker drifting.
The proposed model with cycle consistency has shown to be most robust as masks propagate.}
  \label{fig:abl_frames}
}
\capbtabbox{%
\footnotesize\addtolength{\tabcolsep}{-5pt}
\begin{tabular}{cccc}
\toprule
Method & $\mathcal{J}$(Mean) & $\mathcal{F}$(Mean) \\ \midrule
Ours (Full Model)                         & 47.7   & 51.3  &  \\
Ours w/o Colour Dropout              & 40.5   & 39.5  &  \\
Ours w/o Restricted Attention        & 40.8  &  39.7 &  \\    
Ours w/o Scheduled Sampling    & 40.2  & 39.2  &  \\
Ours w/o Cycle Consistency       & 41.0   & 40.4 &  \\  
\bottomrule
\end{tabular}
}
{%
\caption{Ablation Studies on DAVIS-2017. 
$\mathcal{J}$: region overlapping, $\mathcal{F}$: contour accuracy respectively.}
\label{table:davis_ablation}
}
\end{floatrow}
\end{figure}

\vspace{-6pt}
\par \noindent \textbf{Colour Dropout}
Instead of taking full-colour input, 
we follow Vondrick~\etal~\cite{Vondrick18}, and convert all frames into grayscale for inputs.
As shown in Table~\ref{table:davis_ablation}, 
both metrics drop significantly, \ie $7.2\%$ in $\mathcal{J}$ and $11.8\%$ in $\mathcal{F}$. 
This demonstrates the importance of bridging the discrepancy between training and testing on utilizing \emph{full-RGB} colour information.

\vspace{2pt}
\par \noindent \textbf{Restricted Attention}
When computing the affinity matrix with full attention, 
the model makes use of about $9.2$G of GPU memory to process a single $480$p image, 
it is therefore impossible to train on \emph{high-resolution} frames with large batch size on standard GPUs~(12-24GB).
In comparison, our model with restricted attention only takes $1.4$G GPU memory for the same image.
As Table~\ref{table:davis_ablation} shows, 
performance dropped by $6.9\%$ and $11.6\%$ on $\mathcal{J}$ and $\mathcal{F}$ before or after using restricted attention. 
This decrease confirms our assumption about spatial coherence in videos,
leading both a decrease of memory consumption, 
and an effective regularizer that avoids the model matching correspondences very far away, 
for instance matching repeated patterns along the entire image.

\vspace{2pt}
\par \noindent \textbf{Scheduled Sampling}
When not using the scheduled sampling, \ie all frames used for copying are groundtruth during training, 
the performance dropped significantly from $47.7$ to $40.2$ in $\mathcal{J}$, $51.3$ to $39.2$ in $\mathcal{F}$,
suggesting that the scheduled sampling indeed improves the model robustness under challenging scenarios, \eg illumination change.

\vspace{2pt}
\par \noindent \textbf{Cycle Consistency}
Lastly, we evalute the effectiveness of forward-backward consistency.
As seen in Figure~\ref{fig:abl_frames}, 
while both models start off with high accuracy early in a video sequence, 
the model with cycle-consistency maintains a higher performance in later stages of video sequences, 
indicating a less severe drifting problem. 
This can also be reflected in the quantitative analysis, 
where cycle consistency enables a performance boost by $7.5\%$ in $\mathcal{J}$ and $12.1\%$ in $\mathcal{F}$. 

% --------------------
\subsubsection{Comparison with State-of-the-art}
% ===============
In Table~\ref{table:davis_1}, 
we show comparisons with previous approaches on the DAVIS-2017 video segmentation benchmark. 
Three phenomena can be observed:
\emph{First},
Our model clearly dominates all the self-supervised method,
surpassing both video colourization~($49.5$ vs. $34.0$ on $\mathcal{J}$\&$\mathcal{F}$) 
and CycleTime~($49.5$ vs. $40.7$ on $\mathcal{J}$\&$\mathcal{F}$).
\emph{Second}, 
our model outperforms the optical flow method significantly, 
suggesting the colour dropout and scheduled sampling has improved the model's robustness 
under scenarios that optical flow is deemed to fail, \eg large illunimation changes.
\emph{Third},
our model trained with self-supervised learning can even approach the results of some supervised methods,
for instance, despite of only using a ResNet18 as feature encoder,
the results are comparable with the ResNet50 pretrained on ImageNet~($49.5$ vs. $49.7$ on $\mathcal{J}$\&$\mathcal{F}~(Mean)$).
We conjecture this is due to the fact that the model pretrained with ImageNet has only encoded high-level semantics, 
while not optimized for dense correspondence matching.

% ===============
\begin{table}[!htb]
\centering
\footnotesize\addtolength{\tabcolsep}{-5pt}
\begin{tabular*}{\textwidth}{c @{\extracolsep{\fill}} cccccccc}
\toprule
Method & Supervised & Dataset& $\mathcal{J}$\&$\mathcal{F}$(Mean) & $\mathcal{J}$(Mean) & $\mathcal{J}$(Recall) & $\mathcal{F}$(Mean) & $\mathcal{F}$(Recall) \\ \midrule

Identity              &  \xmark & - &22.9 & 22.1 & 15.9 & 23.6 & 11.7 \\ 
Optical Flow (FlowNet2)~\cite{Ilg17} &  \xmark  & - & 26.0 & 26.7 & - & 25.2 & - \\
SIFT Flow~\cite{Liu11}             &  \xmark  & - & 34.0 & 33.0 & - &  35.0 & - \\
Transitive Inv.~\cite{Wang17}       &  \xmark  & - & 29.4 & 32.0 & - & 26.8 & - \\
DeepCluster~\cite{Xie16}           &  \xmark  & YFCC100M & 35.4 & 37.5 & - & 33.2 & - \\
Video Colorization~\cite{Vondrick18}    & \xmark   & Kinetics & 34.0 & 34.6 & 34.1 & 32.7 & 26.8\\
CycleTime (ResNet-50)~\cite{Wang19} & \xmark   & VLOG & 40.7 & 41.9 & 40.9 & 39.4 & 33.6 \\
\textbf{Ours (Full Model ResNet-18)}                  & \xmark   & Kinetics~\cite{Kay17} & \textbf{49.5} & \textbf{47.7} & \textbf{53.2} & \textbf{51.3} & \textbf{56.5} \\  
\textbf{Ours (Full Model ResNet-18)}                  & \xmark   & OxUvA~\cite{Valmadre18} & \textbf{50.3} & \textbf{48.4} & \textbf{53.2} & \textbf{52.2} & \textbf{56.0} \\  
\midrule
ImageNet (ResNet-50)~\cite{He16} & \cmark  & ImageNet & 49.7 & 50.3 & - & 49.0 & - \\
SiamMask~\cite{Wang19a}   & \cmark  & YouTube-VOS & 53.1 & 51.1 & 60.5 & 55.0 & 64.3\\
OSVOS\cite{caelles2017one}     & \cmark  & DAVIS & 60.3 & 56.6 & 63.8 & 63.9 & 73.8     \\  
\bottomrule
\end{tabular*}
\vspace{-8pt}
\caption{Video segmentation results on DAVIS-2017 dataset. Higher values are better.}
\label{table:davis_1}
\end{table}

% ===============
\vspace{-15pt}
\subsubsection{Accuracy by Attributes}
\begin{figure*}[!htb]
  \centering
  \includegraphics[width=.9\textwidth]{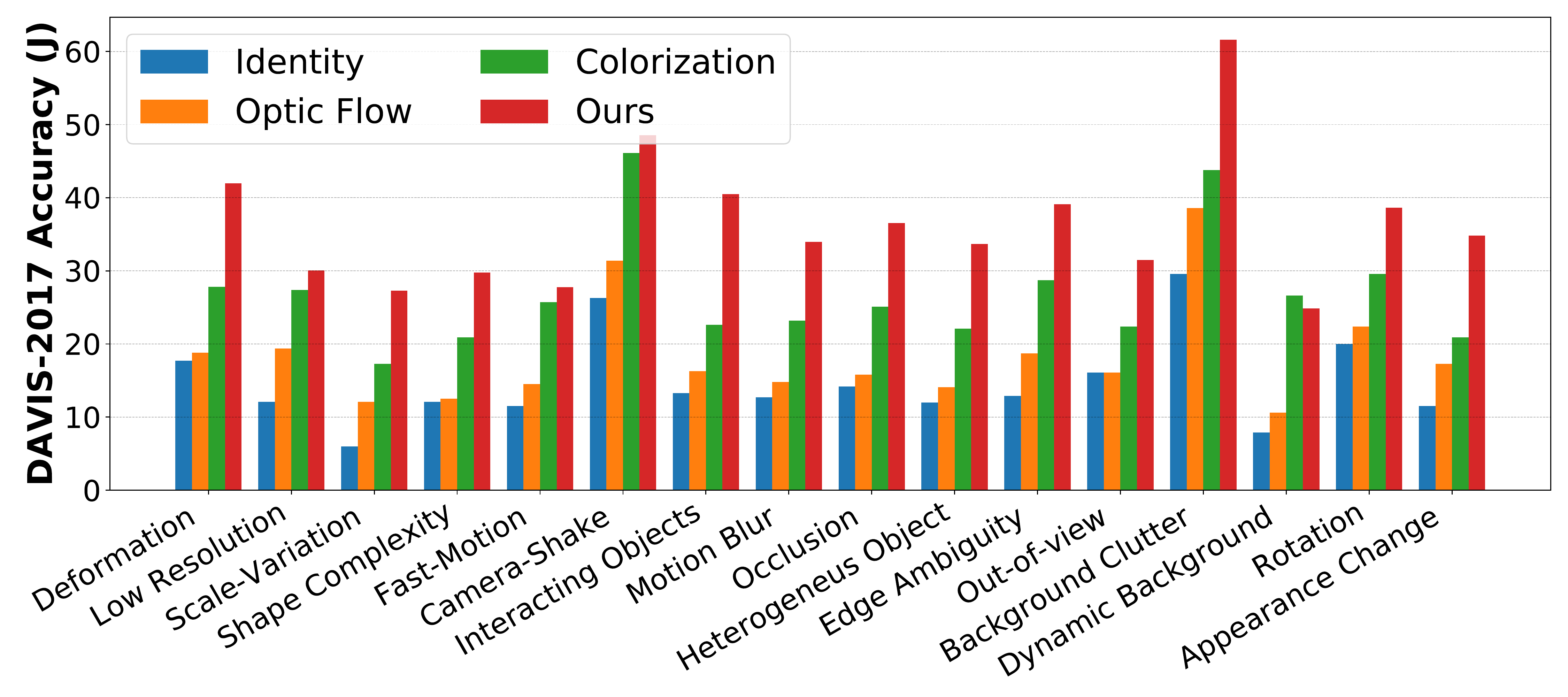}
  \vspace{-10pt}
  \caption{Accuracy by attributes.}
  \label{fig:attributes}
\end{figure*}

In Figure~\ref{fig:attributes}, 
we show DAVIS-2017 testing accuracy grouped into different categories. 
The proposed method has shown to outperform the previous methods in all situations,
suggesting that our model has learned better feature embeddings for robust correspondence flow.

% --------------------
\vspace{-6pt}
\subsubsection{Qualitative Results}
\begin{figure*}[!h]
  \centering
%  % Requires \usepackage{graphicx}
  \includegraphics[width=.9\textwidth]{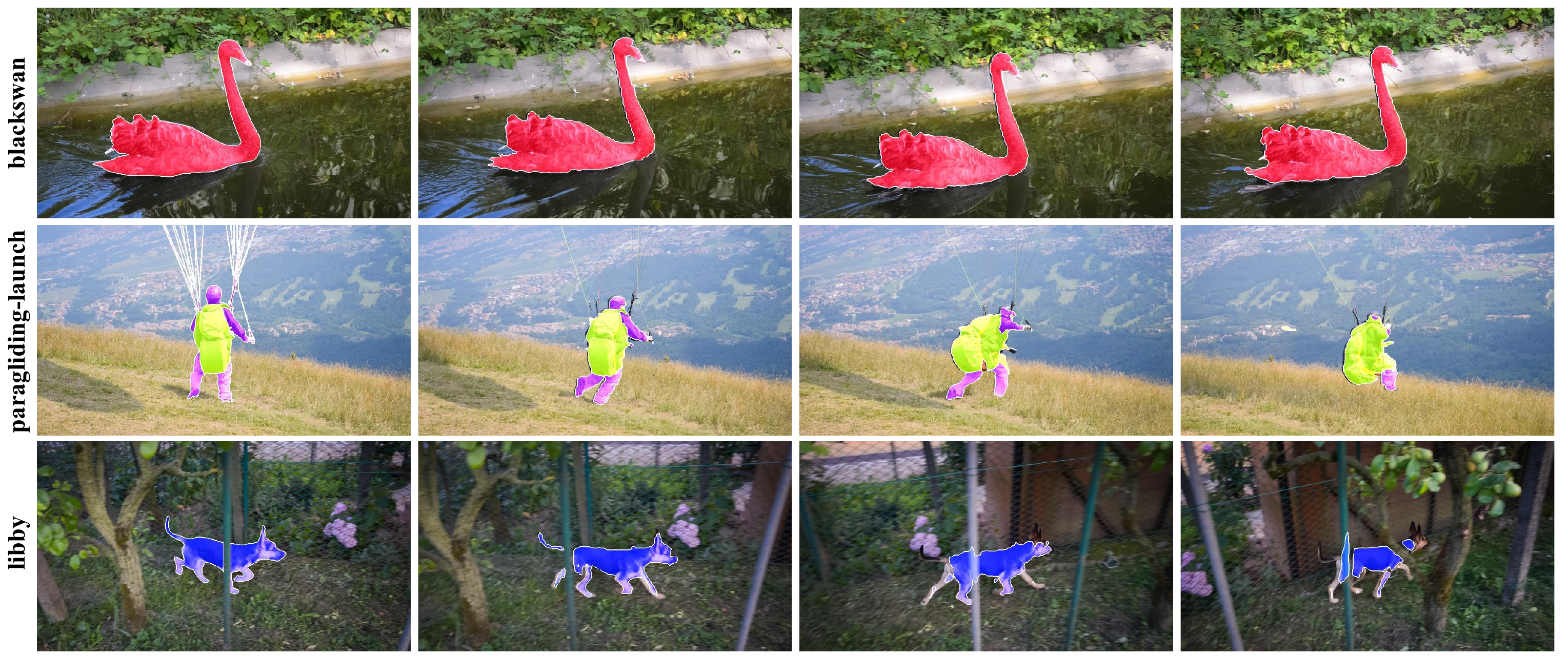}
  \vspace{-10pt}
  \caption{Qualitative results on DAVIS-2017.}
  \label{fig:qualitative_2}
\end{figure*}

% ===============
As shown in Figure~\ref{fig:qualitative} and Figure~\ref{fig:qualitative_2}, 
we provide the qualitative prediction from our model.
The segmentation mask can be propagated through sequences even when facing large scale variation from camera motion, 
and object deformations.

% ===============

% ===============
\vspace{-15pt}
\subsubsection{Probing Upper Bound of Self-supervised Learning}
Despite the superior performance on video segmentation,
we notice that training models on Kinetics is not ideal, as it is a human-centric video dataset.
However, most of the classes in DAVIS are not covered by Kinetics, \eg cars, animals.

To shed light on the potential of self-supervised learning on the task of correspondence flow,
we probe the upper bound by training on more diverse video data. 
We randomly pick $8$ DAVIS classes and download $50$ additional videos from YouTube,
and further train the model by varying the amount of additional data.
Note that, we only download videos by the class labels, 
and \emph{no} segmentation annotations are provided during further self-supervised training.

% ===============
\begin{figure}[!htb]
  \centering
%  % Requires \usepackage{graphicx}
  \includegraphics[width=\textwidth]{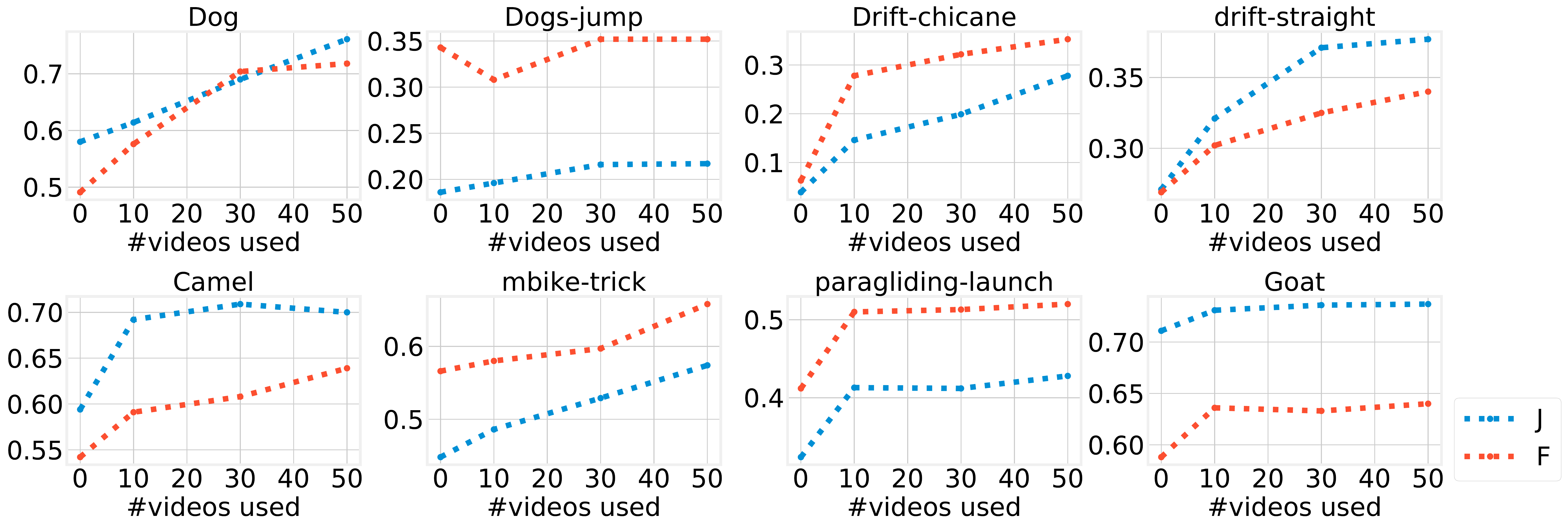}
  \vspace{-25pt}
  \caption{Results after self-supervised learning on additional data.}
  \label{fig:finetune}
\end{figure}

As shown in Figure~\ref{fig:finetune} and Table~\ref{tab:finetune},
two phenomena can be observed:
\emph{First}, 
as the number of additional training videos increases~(Figure~\ref{fig:finetune}), 
all sequences have seen a performance boost on both metrics~($\mathcal{J}, \mathcal{F}$).
\emph{Second}, 
the segmentation on some of the classes are even comparable or surpassing the supervised learning, \eg Drift-c, Camel, Paragliding.
We conjecture this is due to the fact that, 
there are only very limited manual annotations for these classes for supervised learining, \eg Paragliding.
However, with self-supervised learning, we only require raw video data which is almost infinite.

% ===============
\begin{table}[!htb]
\small\addtolength{\tabcolsep}{-2pt}
\footnotesize
\centering
\begin{tabular}{ccccccccccc}
\toprule
 Method & Dataset & Dog & Dog-j & Drift-c & Drift-s & Camel & Mbike & Paragliding & Goat     \\
 \midrule
Self-supervised ($\mathcal{J}$)  & Kinetics  & 58.0 & 18.6 & 3.9 & 27.1 & 59.4 & 44.8  & 32.4 & 71.1  \\
Self-supervised ($\mathcal{J}$) & Additional & 76.1 & 21.7 & 27.8 & 37.7 & 70.0 & 57.4 & 42.8 & 73.7  \\
Supervised ($\mathcal{J}$)~\cite{Yang18}    & COCO+DAVIS  & 87.7 & 38.8 & 4.9 & 66.4 & 88.4 &  72.5 & 38.0 & 80.4  \\
 \midrule
Self-supervised ($\mathcal{F}$)  & Kinetics  & 49.1 & 34.3 & 6.3 & 26.9 & 54.2 & 56.6 & 41.2 & 58.8  \\
Self-supervised ($\mathcal{F}$)  & Additional & 71.8 & 35.2 & 35.3 & 34.0 & 63.9 & 65.8 & 52.0 & 64.0  \\
Supervised ($\mathcal{F}$)~\cite{Yang18}  & COCO+DAVIS     & 84.6 & 45.2 & 8.4 & 57.7 & 92.2 & 76.7 & 58.1 & 74.7  \\
 \bottomrule
\end{tabular}
\vspace{-8pt}
  \caption{Quantitative comparison before and after training on additional videos.}
\label{tab:finetune}
\end{table}

%%%%%% JHMDB %%%%%%
\vspace{-15pt}
\subsection{Keypoint Tracking on JHMDB}
As shown in Table~\ref{table:jhmdb_1}, 
our approach exceeds the previous methods~\cite{Vondrick18} by an average of $11.3\%$ in PCK$_{instance}$.
%suggesting the effectiveness of the proposed ideas, 
%\ie training on \emph{full-colour}, \emph{high-resolution} videos over \emph{long} temporal windows.
Also, we achieve better performance in the more strict PCK@.1 metric when compared to the recent work~\cite{Wang19}. 
Interestingly, 
when taking the benefit of the almost infinite amount of video data,
self-supervised learning methods~(CycleTime and Ours) achieve comparable or even outperforms model trained with the supervised learning\cite{He16, song2017thin}.

\begin{table}[!htb]
\centering
\footnotesize
\begin{tabular*}{\textwidth}{c @{\extracolsep{\fill}} ccccccc}
\toprule
\multirow{2}{*}{Method} & \multirow{2}{*}{Supervised}  &\multirow{2}{*}{Dataset} & \multicolumn{2}{c}{PCK$_{instance}$} & \multicolumn{2}{c}{PCK$_{max}$} \\ 
\cmidrule{4-5} \cmidrule{6-7} 
 &  & & @.1 & @.2 & @.1 & @.2  \\ 
 \midrule
SIFT Flow\cite{Liu11}                                & \xmark  & - & 49.0 & 68.6 & - & - \\ 
Video Colorization~\cite{Vondrick18}     & \xmark & Kinetics  & 45.2 & 69.6 & - & - &  \\ 
CycleTime (ResNet-50)~\cite{Wang19} &\xmark & VLOG  & 57.7 & 78.5 & - & - \\  
\textbf{Ours (Full Model ResNet-18)}     & \xmark & Kinetics & \textbf{58.5} & \textbf{78.8} & \textbf{71.9} & \textbf{88.3} \\  
\midrule
ImageNet (ResNet-50)~\cite{He16}       & \cmark  &ImageNet & 58.4 & 78.4 & - & -  \\  
Fully Supervised~\cite{song2017thin}    &\cmark  & JHMDB & - & - &  68.7 & 81.6 \\  
\bottomrule
\end{tabular*}
\vspace{-10pt}
\caption{Keypoint tracking on JHMDB dataset (validation split 1). Higher values are better.}
\label{table:jhmdb_1}
\end{table}

%!TEX root=../root.tex
\vspace{-20pt}
\section{Conclusion}
\vspace{-5pt}
The paper aims to explore the self-supervised learning for pixel-level correspondence matching in videos.
We proposed a simple \emph{information bottleneck} that enables the model to be trained on standard RGB images, 
and nicely close the gap between training and testing.
To alleviate the challenge from model drifting,
we formulate the model in a recursive manner, trained with scheduled sampling and forward-backward cycle consistency.
We demonstrate state-of-the-art performance on video segmentation and keypoint tracking.
To further shed light on the potential of self-supervised learning on correspondence flow,
we probe the upper bound by training on additional and more diverse video datasets,
and show that self-supervised learning for correspondence flow is far from being saturated.
As future work, potential extensions can be:
\emph{First}, explore better approaches for overcoming tracker drifting, 
\eg use explicit memory modules for long-term correspondence flow.
\emph{Second}, define robust loss functions that can better handle complex object deformation and occlusion, 
\eg predicting visibility mask or apply losses at feature level~\cite{Oord18}.
\emph{Third}, instead of doing quantization with Kmeans clustering, 
train the quantization process to be more semantically meaningful.

\vspace{-10pt}
\subsection*{Acknowledgment}
\vspace{-5pt}
Financial support for this project is provided by EPSRC Seebibyte Grant EP/M013774/1.

\bibliography{shortstrings,vgg_local,vgg_other}

% ------------------------------------------
\clearpage
\begin{appendices}

\section{Network Architecture}
We use a modified ResNet-18\cite{He16} architecture with enlarged output feature maps size. 
Details of the network can be found below.

\begin{table}[h!]
\centering
\scriptsize
\begin{tabular}{l|l}
 \hline
 0 & Input image  \\
 \hline
 \multicolumn{2}{c} { \textbf{Feature extractor} } \\
 \hline
  1 & $7 \!\times\! 7$ conv with stride 2 and 64 filters \\
  2 & $3 \!\times\! 3$ Residual Block with stride 1 and 64 filters \\
  3 & $3 \!\times\! 3$ Residual Block with stride 2 and 128 filters \\
  4 & $3 \!\times\! 3$ Residual Block with stride 1 and 256 filters \\
  5 & $3 \!\times\! 3$ Residual Block with stride 1 and 256 filters \\
\hline
\hline

\end{tabular}
\caption{Network architecture. A Residual Block stands for a residually connected sequence of operations: convolution, batch normalization, rectified linear units (ReLU), convolution, batch normalization. See \cite{He16} for details.
\label{table:network-detail}}
\end{table}

\subsection{Failure Cases}
\begin{figure*}[!htb]
  \centering
%  % Requires \usepackage{graphicx}
  \includegraphics[width=\textwidth]{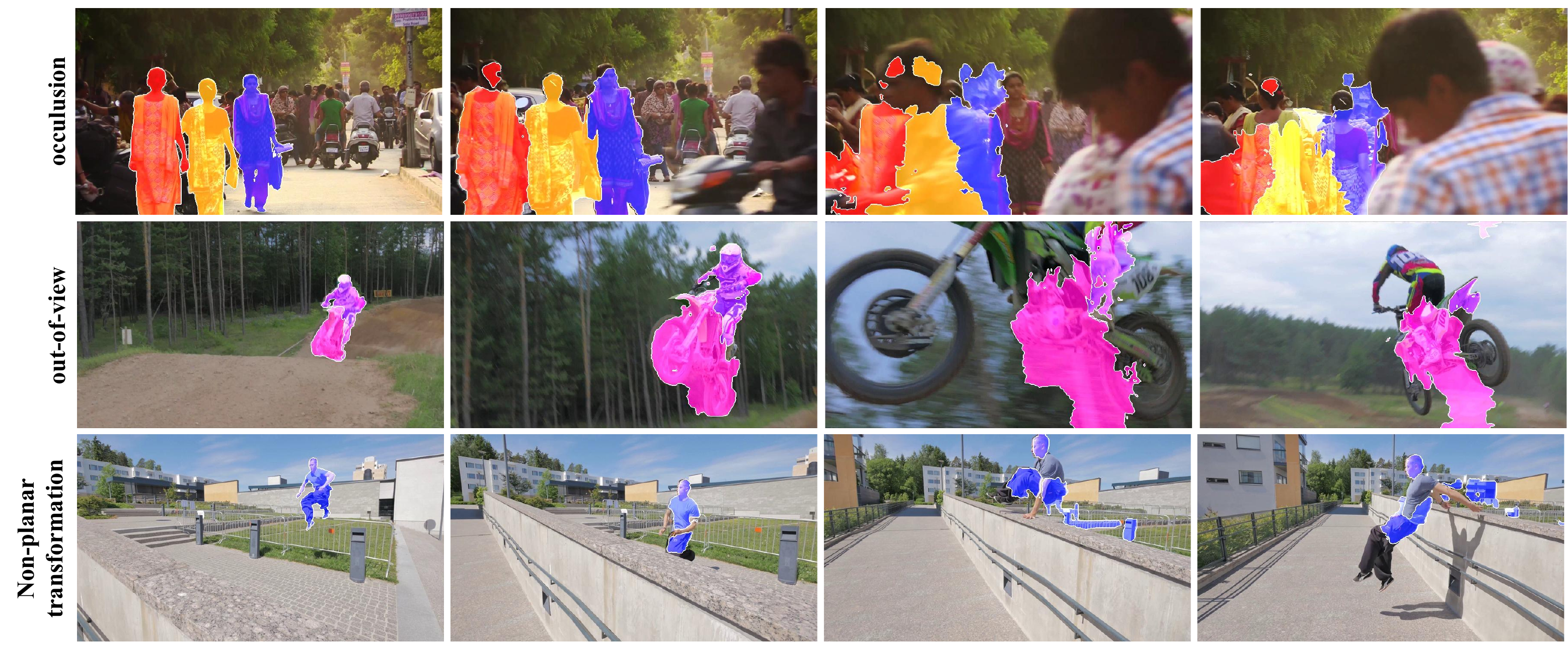}
  \caption{Common failed cases, including occulusion, out-of-view and complex transformation.}
  \label{fig:failure}
\end{figure*}

Figure~\ref{fig:failure} demonstrates some failure cases during mask propagation. 
Row 1 shows our tracker fails due to occlusions. 
As we only use the mask of the previous frame to propagate, 
an object is unlikely to be retrieved after it has been occluded. 
Similarly, it is difficult to recover an object once it goes out of the frame (Row 2). 
Lastly, if the object is under complex deformation,
it is likely to incur the model drifting.

\end{appendices}
% ------------------------------------------

%\vspace{-20pt}
%\section{Additionnal Results on DAVIS-2017}
%\subsection{Accuracy by Attributes}
%In Figure~\ref{fig:attributes_}, 
%we show DAVIS-2017 testing accuracy grouped into different categories. 
%The proposed method has shown to outperform the previous methods in all situations.
%Specifically, on the colour-related classes, \eg Background Clutter, Interactive Objects, 
%our method has gained significant performance boost, 
%suggesting that our model has learned better feature embeddings by using full-colour inputs.
%\begin{figure*}[!htb]
%  \centering
%  % Requires \usepackage{graphicx}
%  \includegraphics[width=\textwidth]{figures/attribute.pdf}
%  \caption{Accuracy by attributes.}
%  \label{fig:attributes_}
%\end{figure*}

% --------------------
%\subsection{Qualitative Results}
% ===============
%As shown in Figure~\ref{fig:qualitative_2}, 
%we provide the qualitative prediction from our correspondence flow model.
%The segmentation mask can be propagated through sequences even when facing large scale variation from camera motion, 
%and object deformations.

% ===============
%\begin{figure*}[!h]
 % \centering
%  % Requires \usepackage{graphicx}
% \includegraphics[width=\textwidth]{figures/qualitative_v2.pdf}
% \vspace{-10pt}
% \caption{Qualitative results on DAVIS-2017.}
% \label{fig:qualitative_2}
%\end{figure*}

\end{document}